# GIMP and Wavelets for Medical Image Processing: Enhancing Images of the Fundus of the Eye

**Amelia Carolina Sparavigna**

Department of Applied Science and Technology, Politecnico di Torino, Torino, Italy

**Abstract:** The visual analysis of retina and of its vascular characteristics is important in the diagnosis and monitoring of diseases of visual perception. In the related medical diagnoses, the digital processing of the fundus images is used to obtain the segmentation of retinal vessels. However, an image segmentation is often requiring methods based on peculiar or complex algorithms: in this paper we will show some alternative approaches obtained by applying freely available tools to enhance, without a specific segmentation, the images of the fundus of the eye. We will see in particular, that combining the use of GIMP, the GNU Image Manipulation Program, with the wavelet filter of Iris, a program well-known for processing astronomical images, the result is giving images which can be alternative of those obtained from segmentation.

**Keywords:** Image processing, Retina, Retina Vessels, GIMP, AstroFracTool, Iris, Wavelets

**1. Introduction**
The visual analysis of retina, the light-sensitive layer of tissue on the inner surface of the eye, is known to be important in diagnosis and monitoring of diseases of visual perception, that can be consequences of hypertension, diabetes and other problems [1,2]. In fact, some eye diseases are coming from problems of the blood vessels in the eye. As discussed in [3], and in several other works (see for instance [4]), the assessment of vascular characteristics plays a fundamental role in the related medical diagnoses. The image processing usually involved in these diagnoses is the "retinal vessel segmentation", and its algorithms are relevant to computer aided retinal disease screening systems [3]. The motivation for these computer aided systems is in avoiding manual delineation of retinal blood vessels that requires extensive training and skill [3,5].
In computer vision, image segmentation is the process of partitioning a digital image into multiple "segments", which are sets of pixels. The aim of this method is that of changing the image into something easier to analyze. Image segmentation is typically used to locate lines, curves and edges in the image frames [6]. The segmentation has many applications in medical imaging: in [3], several references are given about segmentations applied to retinal analysis and a new method is proposed.
The approach to image segmentation can be quite complex. In this paper we will see some alternative methods, based on the use of freely available tools for image processing, to enhance in a simple manner the images of the back of the eye. We will see in particular, that combining the use of a wavelet filter with GIMP, the GNU Image Manipulation Program, the final result produces a segmentation-like effect and images which can be alternative of those obtained from segmentation. The wavelet filter is a tool included in a freely downloadable software (Iris), well-known for the processing of astronomical images.

**2. Fundus images and Databases**
The images of the back of the eye are obtained by the fundus photography, able to give a digital image of the retina, optic disc, and macula [7,8]. Fundus photography is used to diagnose and monitor progression of a disease. It is needed to obtain measurements of vessel width, colour, reflectivity, and so on. Some databases containing digital images are available for studying the problems of fundus. One is DRIVE [9].
According to [9], DRIVE database has been established to enable comparative studies on segmentation of blood vessels in retinal images. The web site is also showing very good segmentations, given as described in [10]. Another database with retinal images in which the vasculature has been segmented is available on Adam Hoover's Stare web site [11].

**3. GIMP, Iris and AstroFracTool**
To process an image, we can use GIMP, the GNU Image Manipulation Program, which is freely available on the Web. This program allows processing the images to increase the visibility of their details, simply adjusting brightness and contrast or using several other methods, available in this program. We used GIMP to analyze satellite images: for instance we investigated the motion of sand dunes with this software [12]. GIMP has also several filters for edge detections: in [13], the Sobel filter was used for the study of the dunes on Mars.
Another image processing program that we used for the analysis of satellite images is Iris, a tool created by Christian Buil, which is a well-known software for the analysis of astronomical images [14]. Iris has many pre-processing and processing filters: among them we find the wavelet filter. Launching the corresponding mask, we have some adjustable resolution levels. Changing interactively the levels, it is possible to enhance all the features, small or large, of the image texture.
To enhance the edges of an image, we can also use AstroFracTool [15]. This is a tool based on the implementation of a fractional differential calculation. The fractional differentiation is modifying the image, enhancing the edges but maintaining the original image. Therefore, it is different from the usual edge detection [16].

**4. Processing of fundus photographs**
Let us start using an image from Ref.17. We show it in the Figure 1.a. In this fundus photograph we can see the blood vessels in a normal human retina. As told in [17], veins are darker and slightly wider than corresponding arteries. The optic disc is at left, and the macula lutea is near the center. The gaze is into the camera, and then the macula is at the center of the image, and the optic disk is located towards the nose, which is at left. This disk has pigmentation at the perimeter which is considered non-pathological. Major nerve pathways are seen as white striped patterns radiating from the optic disk.
In the Figure 1, we see how GIMP can change the original image. We have in 1.a the original image and 1.b the image in grey tones. Images 1.c and 1.d are obtained by adjusting brightness and contrast of 1.a and 1.b. To obtain image 1.e, we applied the GIMP tool Bump Map. With this tool, edges and details appear to be raised. In 1.f, we used the GIMP Sobel filter for edge detection (more details on these tools are given by the GIMP tutorial).
In the Figure 2, we are showing the images obtained using AstroFracTool and Iris, applied to the image 1.a. In 2.a, we are showing the output of AstroFracTool with parameters $v=0.7$ and $\alpha=0.7$ (see Ref.15 for details); in 2.b, image 2.a is enhanced adjusting brightness and contrast with GIMP. In 2.c, we have the output of AstroFracTool with parameters $v=0.9$ and $\alpha=0.5$. In 2.d, image 2.c is enhanced with GIMP. In 2.e, we have the output of Iris wavelet filter (see Reference [18] for details) . The parameter of the resolution "fine" is set at level 10.1, the other parameters are at level 1. In 2.f, we have the resolution levels "finest", "fine" and "medium" at level 10.1, the others equal to 1.
In Figure 3, we are investigating the effect of the background on the output of proposed filters. Instead of the black background which we see in the original image, we use a colour tone of the fundus . The new image is 3.a. In 3.b, we see the image obtained applying GIMP tool Bump Map. Image 3.c is obtained using the GIMP Sobel filter, and in 3.d this output is enhanced adjusting brightness and contrast. In 3.e, we have the AstroFracTool output with parameters

ν=0.7 and α=0.7. In 3.f, we see the output of Iris with resolution levels "finest", "fine" and "medium" at value 10.1, the others equal to 1. Note that the change of the background is influencing the final result.
In Figure 4, we are processing an image from the DRIVE database [9]. We have the original image 4.a; in the image 4.b, the output of the GIMP Bump Map filter and in 4.c that after using the GIMP Sobel filter. In 4.d, we change the background. In 4.e, we see the output of AstroFracTool with parameters ν=0.7 and α=0.7, adjusted with GIMP. In 4.f, we have the output of Iris, levels "finest", "fine" and "medium" at value 10.1, the others equal to level 1.

## 5. The GIMP Cartoon filter
GIMP has several artistic filters. One of them is the Cartoon filter. It modifies the image so that it looks like a cartoon drawing. As told by the tutorial, "its result is similar to a black felt pen drawing subsequently shaded with color. This is achieved by darkening areas that are already distinctly darker than their neighborhood". In the framework of the application we are discussing, it seems than that this tool is able to substitute a manual delineation of retinal blood vessels. In Figure 5, we used the GIMP Cartoon filter. In 5.a and 5.b, there are the original images; in 5.a1 and 5.b1, the two images obtained using Cartoon and in 5.a2 and 5.b2 these two images after an adjustment of brightness and contrast.

## 6. Wavelets
Let us compare the images in Figure 5 with those obtained from another approach based on wavelets. In Figure 6, we have again the original images 6.a and 6.b: in images 6.a1 and 6.b1 we have images obtained after the use of Iris wavelet filter, with resolution levels "finest", "fine" and "medium" at value 25, the level of "Remain" at 2, and the other levels equal to 1. In images 6.a2 and 6.b2, after the wavelet filter, Bump Map of GIMP had been used. The results that we have in the Figure 6 are quite good.

## 7. Use the grey tones
After Figure 5, it seems that GIMP Cartoon is able to solve in a simple way the problem of detecting the blood vessels. But this is not so, because on some images of the DRIVE database, it is not working well. In fact, for these images, we can use the approach based on wavelets, such as that used for Figure 6. However, we can see that GIMP Cartoon can be used on the corresponding grey tone images: the result is shown in Figure 7. In this figure, in the panel 7.a1 we see the original image from DRIVE (img001.ppm). In 7.a2 we have the corresponding grey tone image and in 7.a3 the image we obtain using GIMP Cartoon filter. In image 7.a4, we have the result of Iris wavelet filter with parameters as in Figure 6, and in the following images 7.a5 and 7.a6, what we obtain after adjusting brightness and contrast and after using the Bump Map. We repeat the same for another image of DRIVE database (see Figure 8).

## 8. Conclusion
All the methods proposed in this paper are able to enhance the images of the fundus of the eye. What could be the best approach for diagnosis and monitoring of diseases is a problem that could be solved by their use in medical diagnoses. To the author of this paper, an approach based on grey tone images and GIMP Cartoon seems simple and easy to use (a set of results from img003-img008 of DRIVE is shown in Figure 9). However, the use of wavelet filters on the colour images seems able to give more details of retinal vessels (Figure 10). In the images we are showing, we used for all them the same levels for wavelets parameters. Of course, these levels can be adjusted to have the best result for each image.

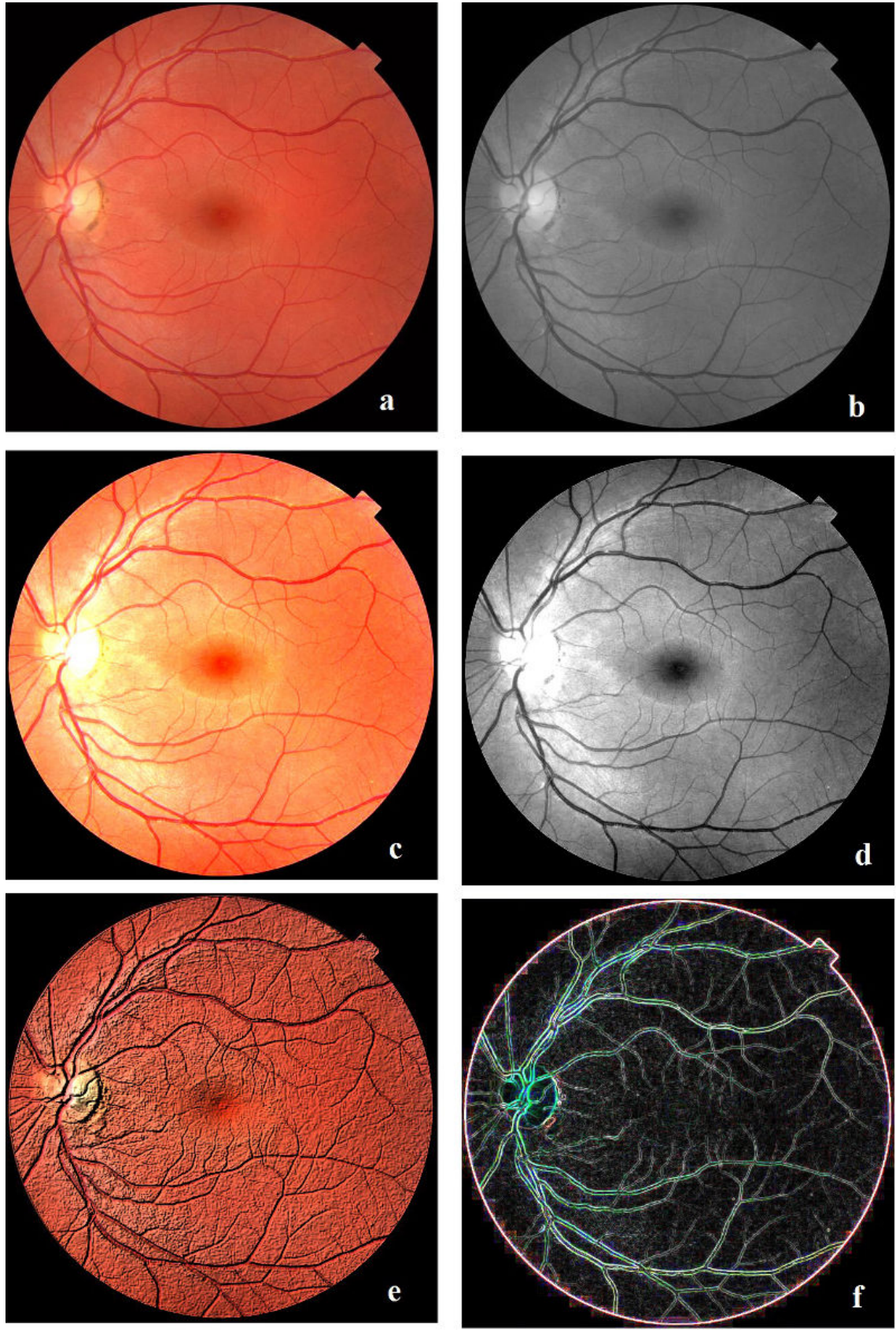

**Figure 1** – 1.a is showing the original image from Ref.17; 1.b is the same image in grey tones. 1.c and 1.d are obtained by adjusting brightness and contrast of 1.a and 1.b using GIMP. To obtain the image 1.e, we applied GIMP tool Bump Map. In 1.f, we used the GIMP Sobel filter for edge detection.

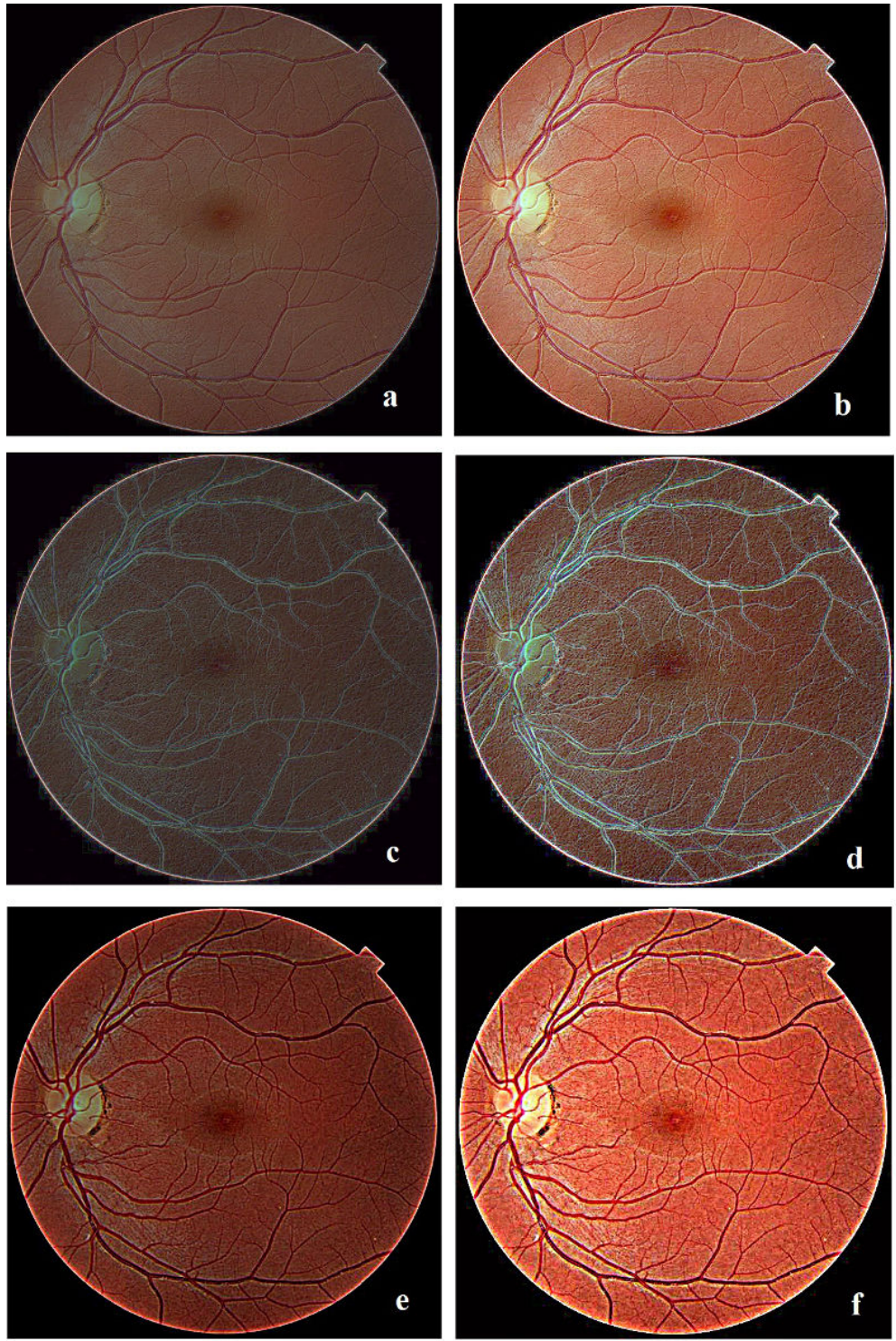

**Figure 2** - 2.a is showing the output of AstroFracTool with parameters ν=0.7 and α=0.7 (see Ref.15 for details); in 2.b, image 2.a is enhanced adjusting brightness and contrast using GIMP. In 2.c, we have the output of AstroFracTool with parameters ν=0.9 and α=0.5. In 2.d, image 2.c is enhanced. In 2.e, we have the output of Iris wavelet filter (see Reference [18] for details) . The parameter of resolution "fine" is set at level 10.1, the other parameters are at level 1. In 2.f, we have resolutions "finest", "fine" and "medium" at level 10.1, the others at level 1.

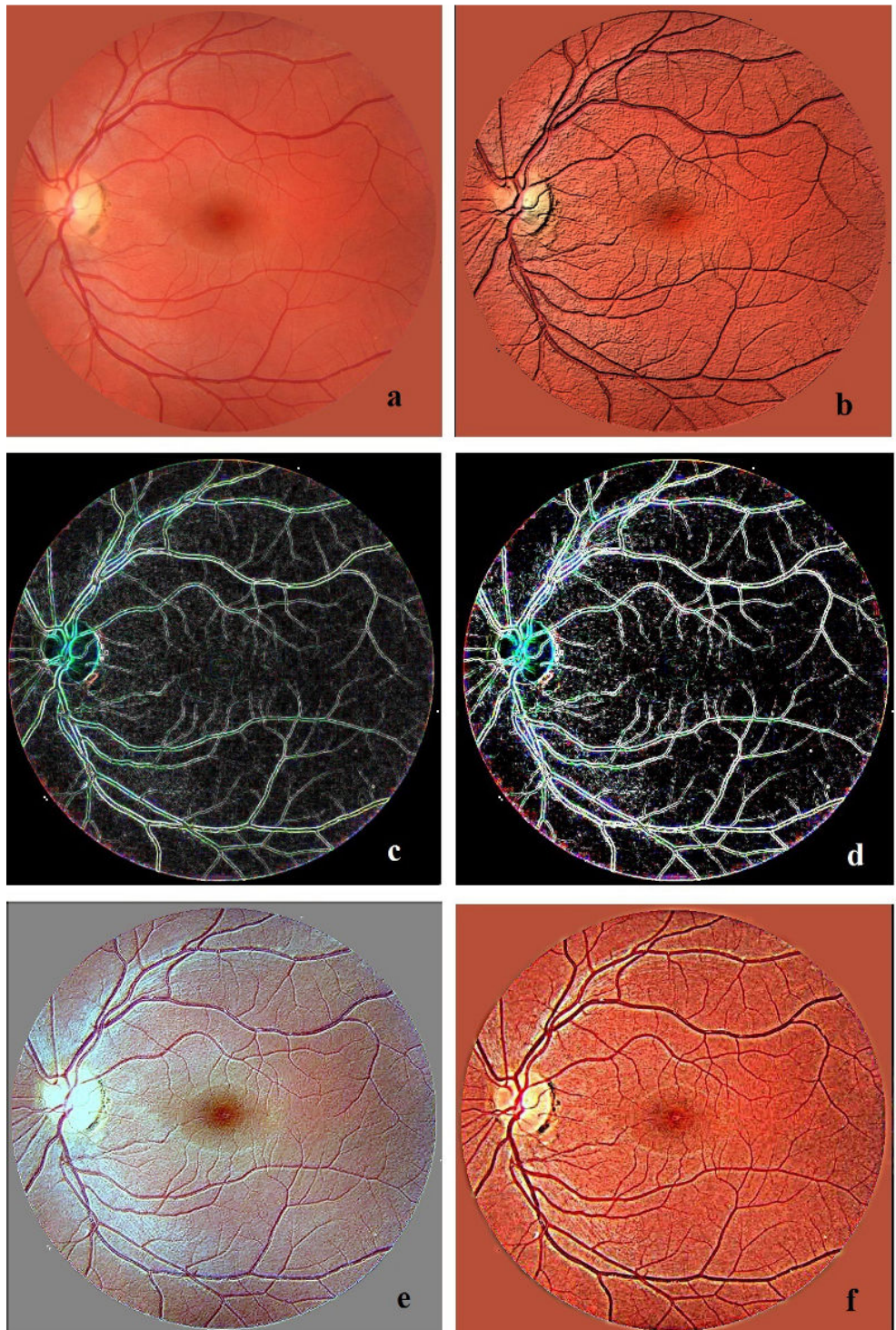

**Figure 3** – Here we are investigating the effect of the background on output images. Instead of a black background, we use a colour tone of the original image. The new image is 3.a. In 3.b, we see the image obtained applying GIMP tool Bump Map. 3.c is obtained using the GIMP Sobel filter, and in 3.d this output is enhances adjusting brightness and contrast. In 3.e we have the AstroFracTool output with parameters $\nu=0.7$ and $\alpha=0.7$. In 3.f we see the output of Iris with resolution levels "finest", "fine" and "medium" at value 10.1, the others at level 1. Note that the change of the background is influencing the final result.

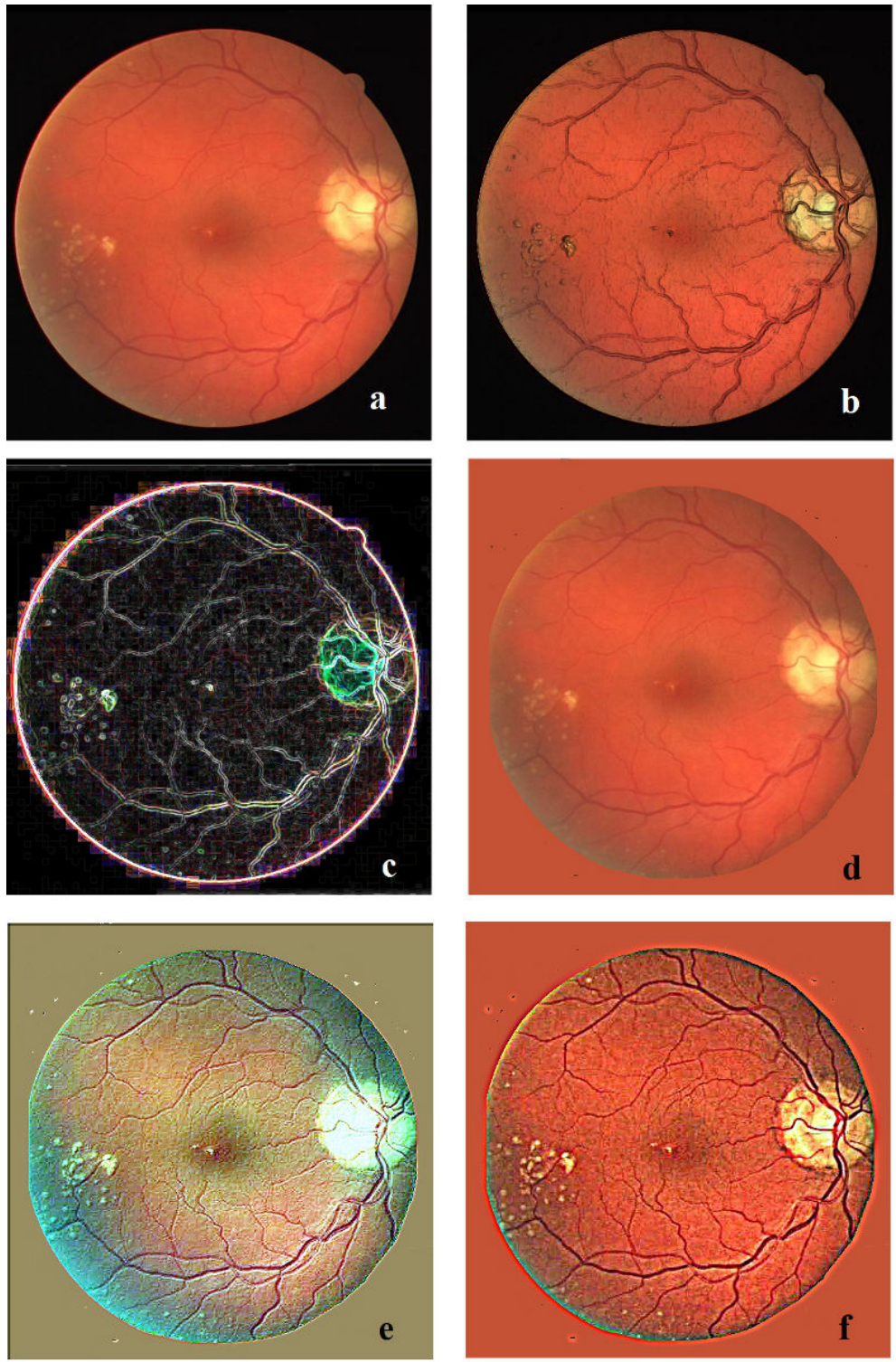

**Figure 4** – Here the processing of an image from DRIVE database [9]. The original image is 4.a; in 4.b, we have the output of GIMP Bump Map filter and in 4.c the image after using the GIMP Sobel filter. In 4.d, we change the background. In 4.e, we see the output of AstroFracTool with parameters $v=0.7$ and $α=0.7$, adjusted with GIMP. In 4.f, we have the output of Iris, levels "finest", "fine" and "medium" at value 10.1, the others equal to 1.

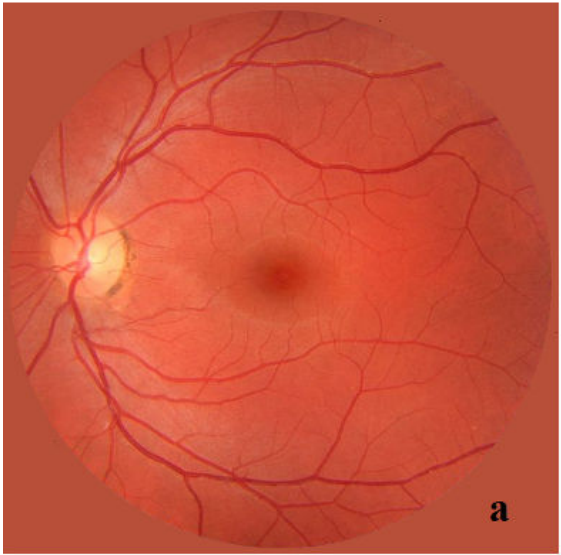
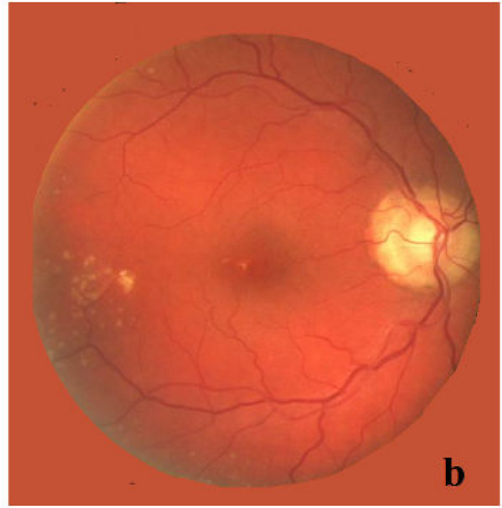
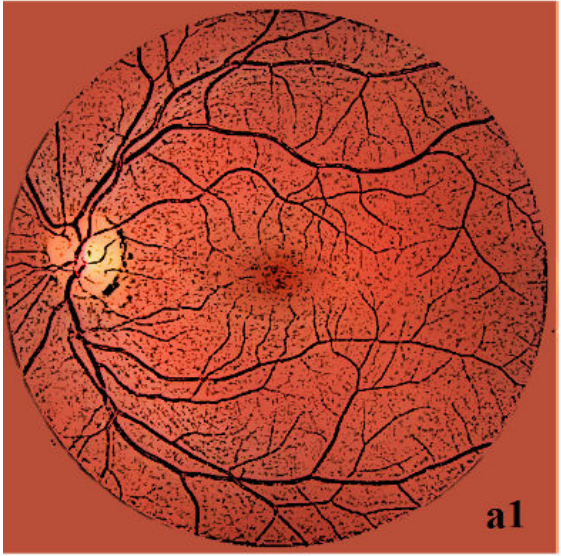
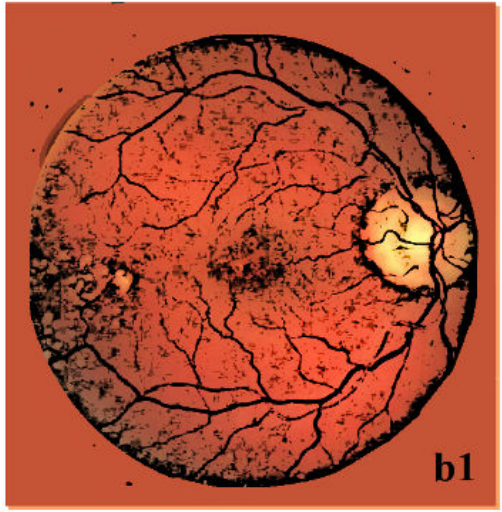
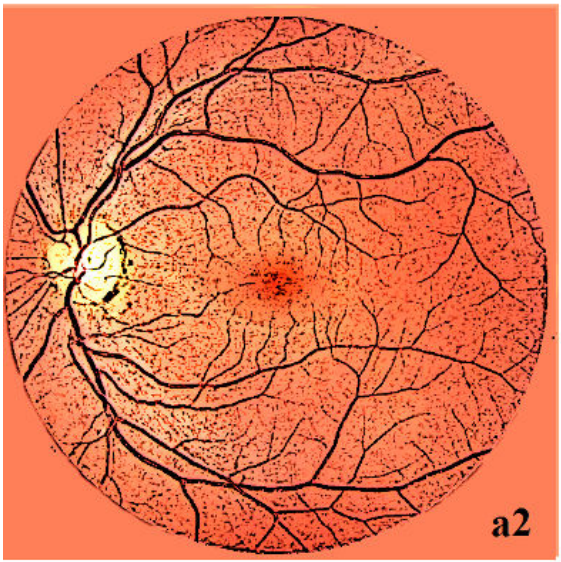
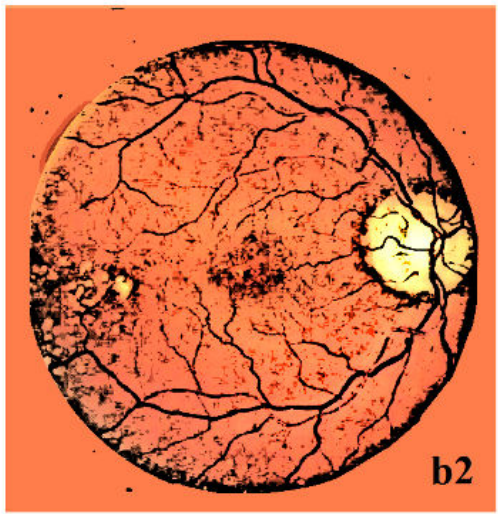

**Figure 5** – GIMP Cartoon filtering. In 5.a and 5.b, the original images. In 5.a1 and 5.b1, the two images obtained using Cartoon and, in 5.a2 and 5.b2, the images after an adjustment of brightness and contrast.

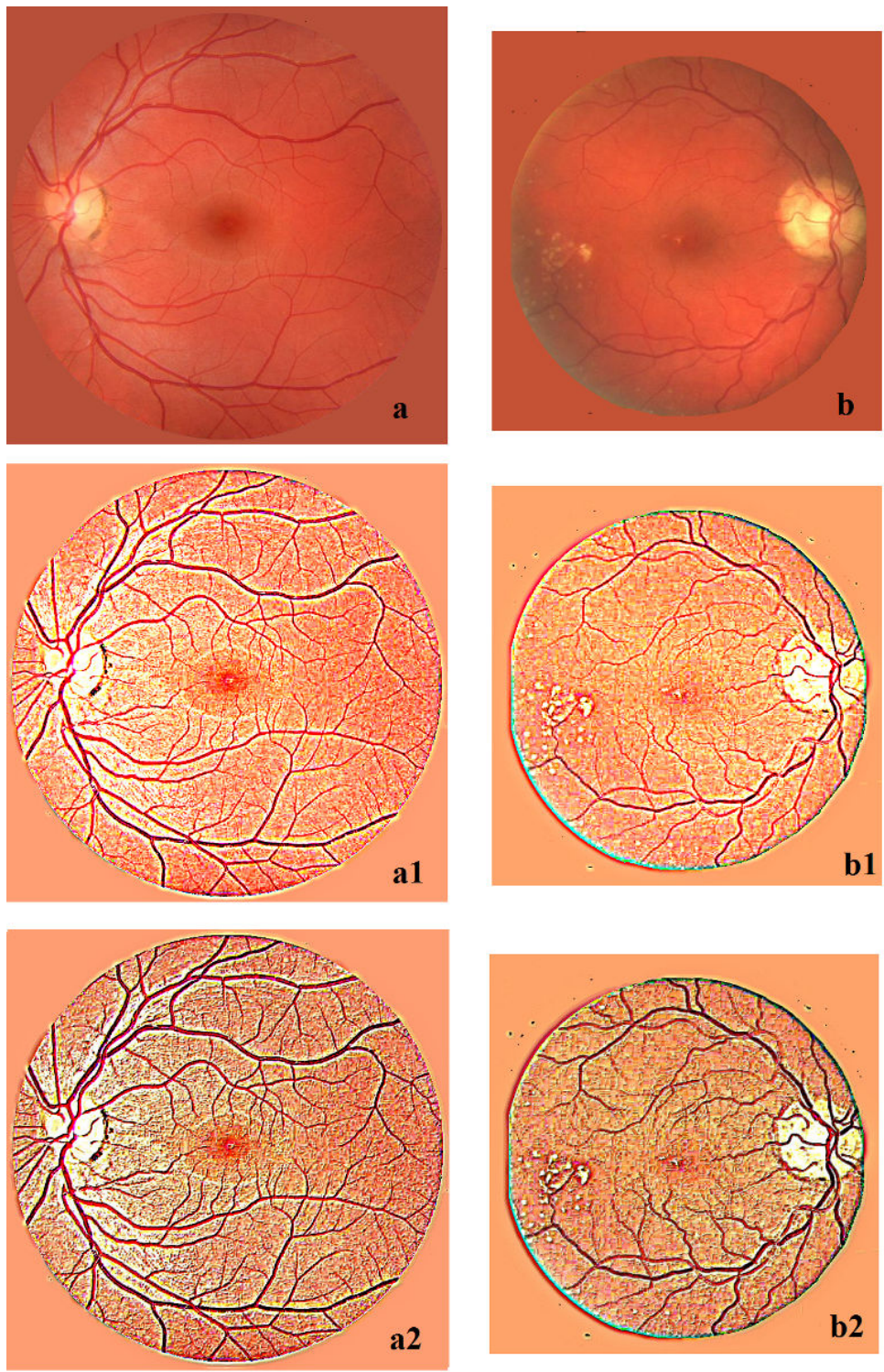

**Figure 6** - Original images are 6.a and 6.b. In images 6.a1 and 6.b1, we have the results obtained using the wavelet filter of Iris, with resolution levels "finest", "fine" and "medium" at value 25, the level of "Remain" at 2, and the others equal to 1. In the images 6.a2 and 6.b2, after the wavelet filter, the Bump Map of GIMP had been used.

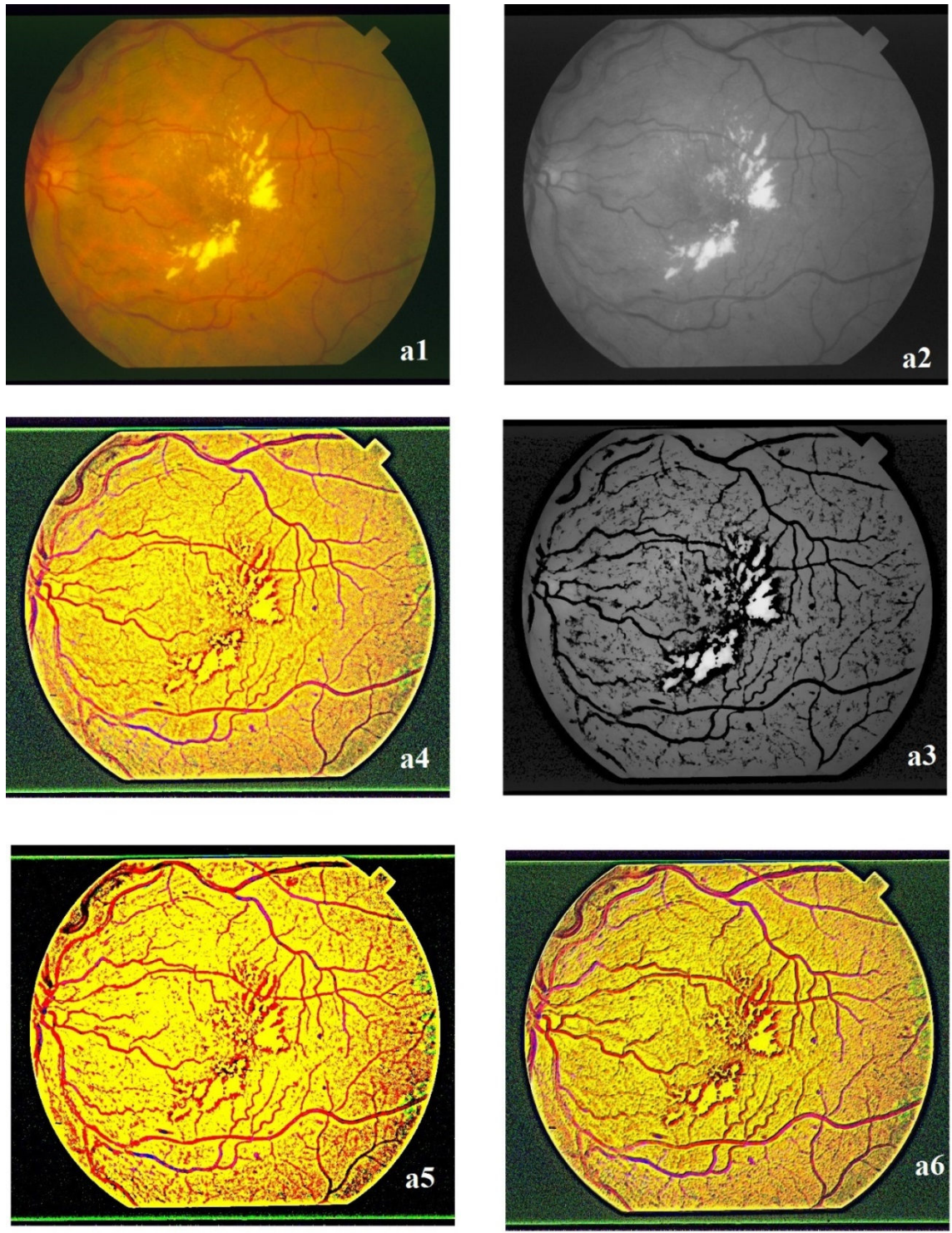

**Figure 7** – In the panel 7.a1, we see the original image from DRIVE (img001.ppm). In 7.a2, we have the corresponding grey tone image and, in 7.a3, the image obtained using GIMP Cartoon filter. In image 7.a4, we have the result of the Iris wavelet filter with parameters like in Figure 6; in images 7.a5 and 7.a6, we see 7.a4 after adjusting brightness and contrast and after using the Bump Map.

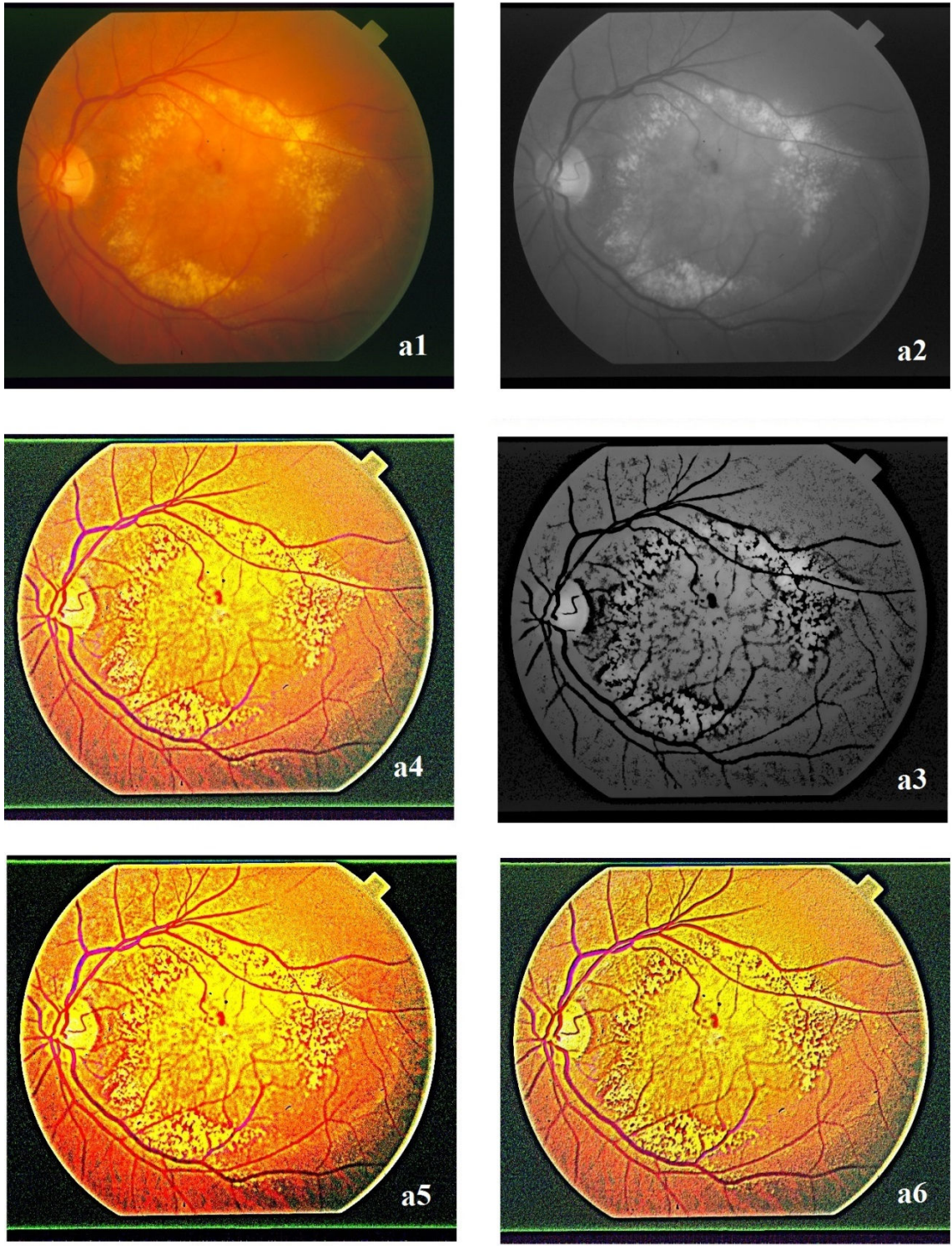

**Figure 8** – In panel 8.a1, we see the original image from DRIVE (img002.ppm). In 8.a2, we have the corresponding grey tone image and in 8.a3 the image we obtain using GIMP Cartoon filter. In image 8.a4, we have the result of Iris wavelet filter, and in the following images 8.a5 and 8.a6, we see it after adjusting brightness and contrast and using Bump Map.

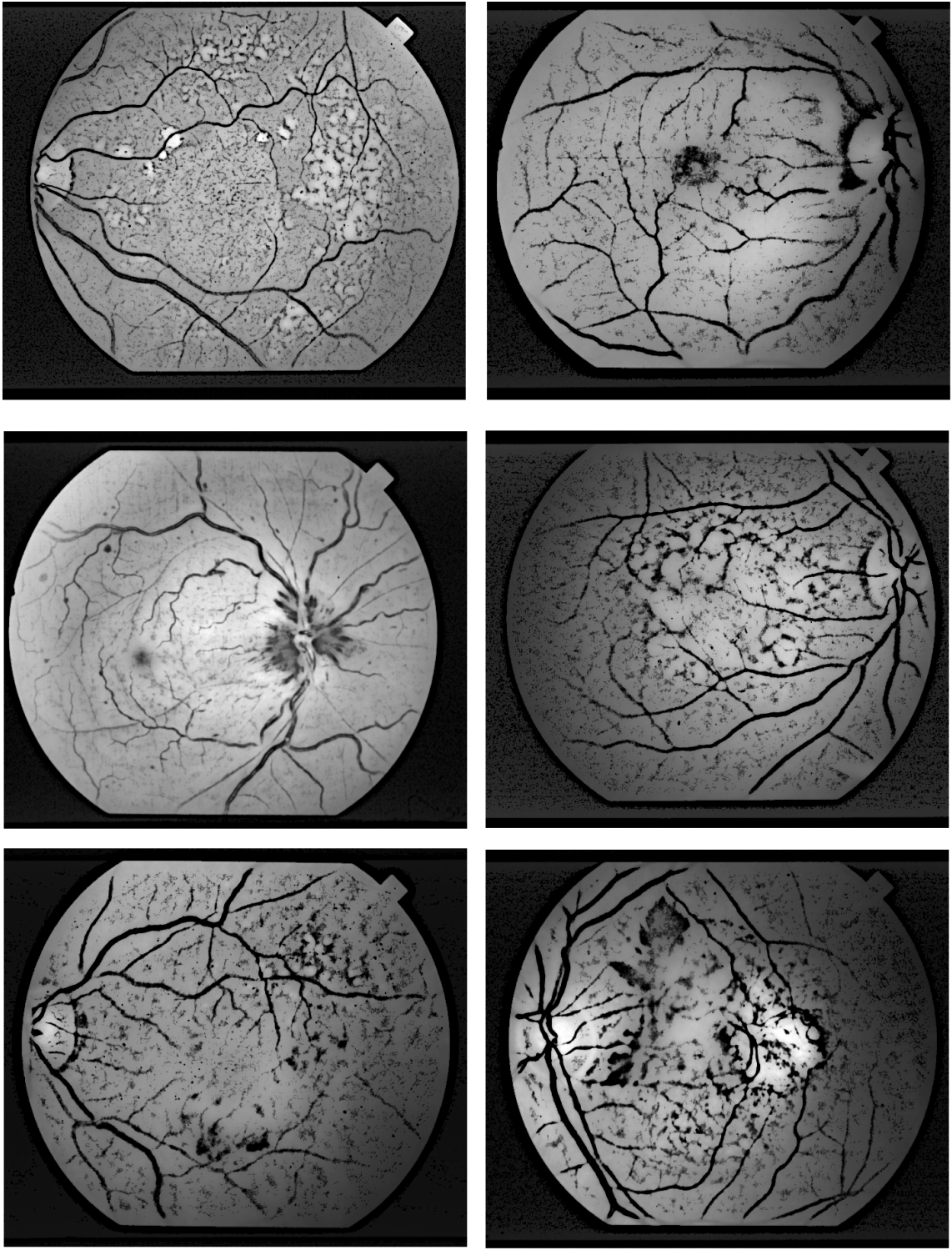

**Figure 9** – The images are obtained using GIMP Cartoon on the grey tone ones of original images (img003-img008.ppm of DRIVE).

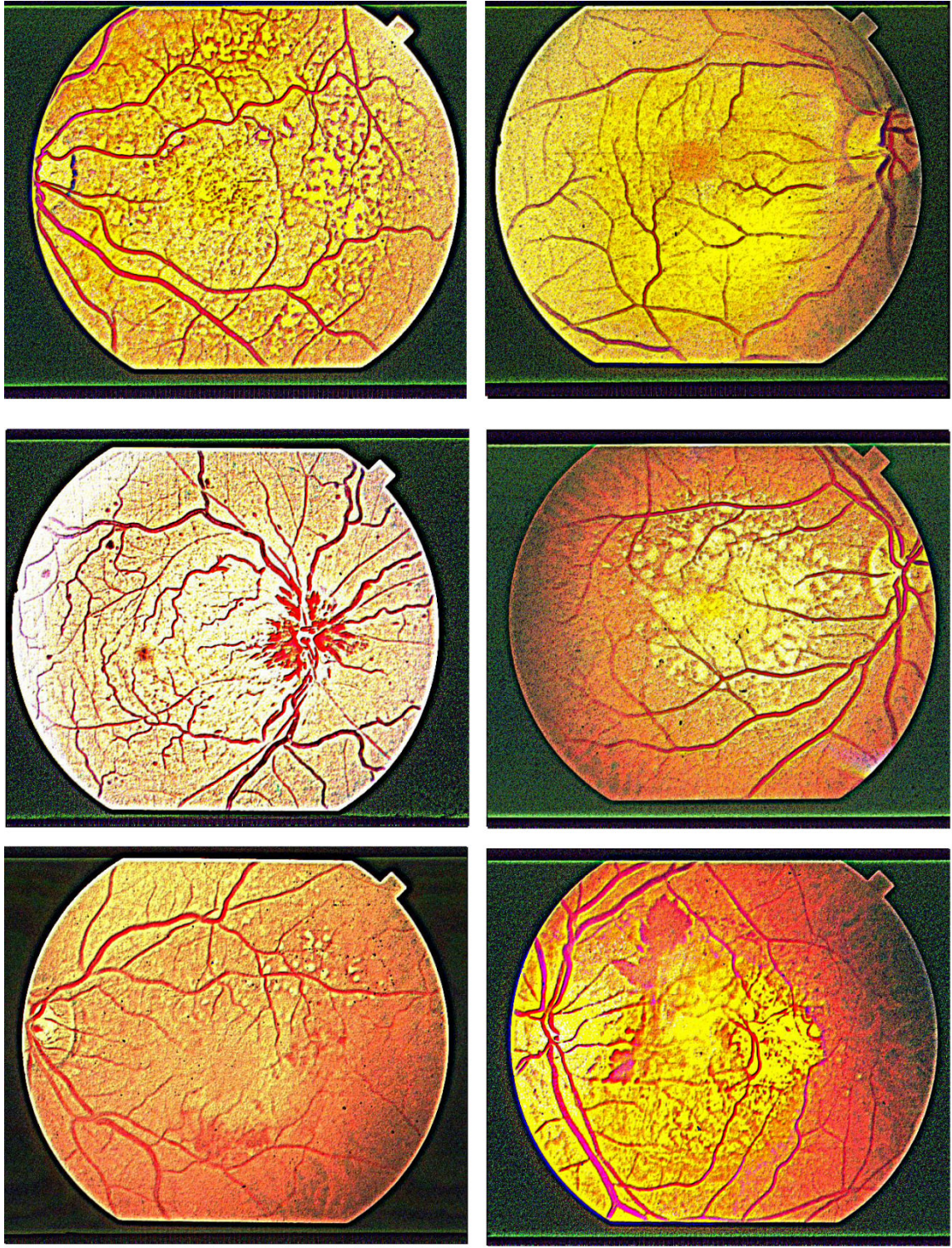

**Figure 10** - The images are obtained using Iris and GIMP on the original images (img003-img008.ppm of DRIVE). The wavelet filter has resolution levels "finest", "fine" and "medium" at value 25, the level of "Remain" at 2, and the others equal to level 1. The output image from Iris was adjusted using the Bump Map of GIMP.